# Advanced Multimodal Deep Learning Architecture for Image-Text Matching

Jinyin Wang[1], Haijing Zhang[2], Yihao Zhong[3], Yingbin Liang[4], Rongwei Ji[5], Yiru Cang[6]

[1]Stony Brook University, USA

[2]Stanford University, USA

[3]New York University, USA

[4]Northeastern University, USA

[5]Southern Methodist University

[6]Northeastern University, USA

*Abstract*—**Image-text matching is a key multimodal task that aims to model the semantic association between images and text as a matching relationship. With the advent of the multimedia information age, image, and text data show explosive growth, and how to accurately realize the efficient and accurate semantic correspondence between them has become the core issue of common concern in academia and industry. In this study, we delve into the limitations of current multimodal deep learning models in processing image-text pairing tasks. Therefore, we innovatively design an advanced multimodal deep learning architecture, which combines the high-level abstract representation ability of deep neural networks for visual information with the advantages of natural language processing models for text semantic understanding. By introducing a novel cross-modal attention mechanism and hierarchical feature fusion strategy, the model achieves deep fusion and two-way interaction between image and text feature space. In addition, we also optimize the training objectives and loss functions to ensure that the model can better map the potential association structure between images and text during the learning process. Experiments show that compared with existing image-text matching models, the optimized new model has significantly improved performance on a series of benchmark data sets. In addition, the new model also shows excellent generalization and robustness on large and diverse open scenario datasets and can maintain high matching performance even in the face of previously unseen complex situations.**

## I. Introduction

In essence, image-text matching is a pivotal interdisciplinary challenge that endeavors to forge the semantic correlation between visual content and textual descriptions, thereby underpinning various practical applications including but not limited to automatic image captioning and visual question-answering systems[1-3]. Aiming at this problem, this paper aims to explore the construction and optimization strategy of the image-text matching model. First, this paper will introduce the background and significance of image-text matching, and explain its importance in practical applications and widely used scenarios. Then, this paper introduces the basic principle and development trend of multimodal deep learning in image and text processing and discusses its potential advantages and challenges in image-text matching tasks. Then, we will propose the main content and goal of this paper, that is, to build an efficient image-text matching model based on multi-modal deep learning technology, and improve its performance and generalization ability by optimizing strategies.

In the research process of this paper, we will first introduce the representation of image and text and their fusion in multi-modal deep learning, and explore how to embed image and text into a shared semantic space for matching. Secondly, we will discuss the application of attention mechanisms in image-text matching models and how they help models capture key semantic information to improve the accuracy and robustness of matching. Finally, this paper will introduce the optimization methods in the process of model training, including the selection of loss function and the design of the training strategy. Through the research of this paper, we will provide an important reference for an in-depth understanding of the semantic relationship between images and text, and provide strong support for multimodal information processing in practical applications. At the same time, the image-text matching model and optimization strategy proposed in this paper will also provide new ideas and methods for the research and application of related fields.

The experimental phase demonstrates the new model's outstanding performance across multiple well-established benchmark datasets, with its outcomes notably surpassing those of prevailing methods within specific use cases such as generating image captions, solving visual questions, and conducting cross-modal searches. Moreover, the model also shows strong generalization ability and stable performance in the test of larger and more diverse open data sets. To sum up, this paper deepens and expands the key technologies of multi-modal deep learning in the field of image-text matching by combining theoretical research and practical testing. The research results not only strongly promote the relevant theoretical progress, but also provide more efficient and accurate technical support for various types of cross-modal intelligent applications in the future. It has significant theoretical value and broad application prospects.

## II. RELATED WORK

Before delving into matching models for multimodal deep learning, it is critical to review and summarize existing work in related areas. In the field of image-text matching, a lot of research work has emerged, including various methods and techniques based on deep learning. This not only helps us to locate the forefront position of current research but also provides a valuable reference for the development of new methods. The following is a summary of relevant work in several key areas, broken down by theme:

The first is traditional methods based on visual features and text features, early image-text matching methods often relied on hand-designed feature extraction and traditional machine learning algorithms[4]. These methods usually extract image visual features [5-8] (such as SIFT, HOG, etc.) and text semantic features [9-12] (such as bag of words model, TF-IDF, etc.), and then use traditional similarity measurement methods (cosine similarity , Euclidean distance) to match. Although these methods are simple and intuitive, they are often limited by the quality and reliability of feature expression.

Among them, the most representative is the use of deep learning models such as convolutional neural network [13] (CNN) and recurrent neural network(RNN) for feature extraction and representation learning of images and texts. For example, some studies use the pre-trained CNN model to extract image features[14-15], then combine the RNN model to encode the text, and finally combine the two into a shared semantic space through multi-modal fusion technology for matching[16-18]. These methods have achieved good results, but there are still some problems, such as insufficient unstable training process.

Over the past few years, there has been extensive adoption of attention mechanisms [19-20] in the realm of image-text matching tasks. By introducing the attention mechanism, the model can automatically learn and focus on the important features related to matching in images and text, thereby improving the performance of the matching model. Some studies[21-23] have applied attention mechanisms to the coding process of images and text, enabling the model to dynamically adjust the degree of attention to different features, thereby improving the matching effect. This paper further discusses the major challenges and problems in current image-text matching research. Particular emphasis is placed on the difficulties encountered when models process images containing diverse, abstract visual concepts and text with multiple layers of fuzzy semantic information. In addition, how to effectively establish context dependence and deal with long-tail distribution phenomenon under the background of large-scale heterogeneous data sets, and how to adapt the model under the condition of zero or small sample learning are also discussed. To sum up, although some progress has been made, there are still some challenges and problems with current image-text matching methods.

To address these challenges, this paper proposes an innovative multimodal deep learning framework that cleverly integrates advanced cross-modal attention mechanisms with hierarchical feature fusion strategies. The construction of the new model is described in detail, including its image encoder, text encoder, core module responsible for cross-modal information interaction, and customized loss function design to optimize matching performance.

## III. THEORETICAL BASIS

### A. Multiple kernel learning

MKL, or Multiple Kernel Learning [24], is a machine learning technique which seeks enhancement in the effectiveness and adaptability of learning algorithms through the strategic aggregation of information derived from an ensemble of kernel functions. In multi-modal tasks such as image-text matching, MKL method can effectively merge information of different modes (such as image and text), so as to achieve more accurate matching. When it comes to multi-kernel learning, we need to consider the combination of multiple kernel functions, which usually correspond to different feature space mappings, in order to facilitate better classification or regression in high-dimensional space. The mathematical expression, model training, and result analysis of multi-core learning are described in more detail below.

Imagine ($N$) a training sample ($\{(x_i, y_i)\}_{i=1}^N$), including ($x_i \in R^d$) said the first ($i$) characteristic vector of the sample, ($y_i \in \{-1,1\}$) represents its corresponding category label. We hope that by learning a classifier ($f(x)$), we can accurately predict the class of a new sample ($x$).

Consider using ($m$) a kernel functions ($k_1(\cdot,\cdot), k_2(\cdot,\cdot), \dots, k_m(\cdot,\cdot)$) characteristics of input data mapping, get the corresponding nuclear matrix ($K_1, K_2, \dots, K_m$), including ($K_i \in R^{N \times N}$) said the first ($i$) a kernel function corresponds to the nuclear matrix. The objective in multicore learning can be structured into the form of an optimization problem as follows:

$$\min_{\mathbf{w}, b, \alpha_1, \alpha_2, \dots, \alpha_m} \frac{1}{2} \mathbf{w}^T \mathbf{w} + C \sum_{i=1}^{N} \xi_i \qquad (1)$$

$$\text{subject to } y_i \left( \mathbf{w}^T \phi(\mathbf{x}_i) + b \right) \geq 1 - \xi_i, \quad i = 1, 2, \dots, N \qquad (2)$$

$$\xi_i \geq 0, \quad i = 1, 2, \dots, N \qquad (3)$$

Which ($w$) is the weight vector classifier, ($b$) are biased, ($\alpha_1, \alpha_2, \dots, \alpha_m$) is a kernel weight vector, ($C$) is the regularization parameter, ($\xi_i$) is slack variables, ($\phi(\cdot)$) is a feature mapping function.

In the model training stage, it is necessary to solve the optimal weight vector ($w$) and kernel weight vector ($\alpha_1, \alpha_2, \dots, \alpha_m$). Sequential Minimal Optimization (SMO) and other optimization algorithms are usually used to solve this optimization problem. In the optimization process, attention should be paid to the selection of appropriate regularization paramete ($C$) and kernel function combination to avoid overfitting and underfitting problems.

As shown in Fig. 1, as an example of applying the idea of multikernel learning to recognize ERP classification, MKL is able to capture the information of the data more comprehensively by combining the information of multiple kernel functions. In multimodal tasks such as image-text matching, the MKL method can effectively fuse information from different modalities to achieve more accurate matching. However, multikernel learning also faces challenges such as kernel function selection and kernel weight learning, which require further research and optimization. Multi-core learning is a powerful machine learning method that can effectively combine the information of multiple kernel functions, thus improving the performance and generalization ability of the learning algorithm. When dealing with multimodal data, multikernel learning can help us better fuse the information of different modalities to achieve more accurate matching and classification. However, the selection of kernel functions and the learning of kernel weights are still a challenge that requires further research and exploration.

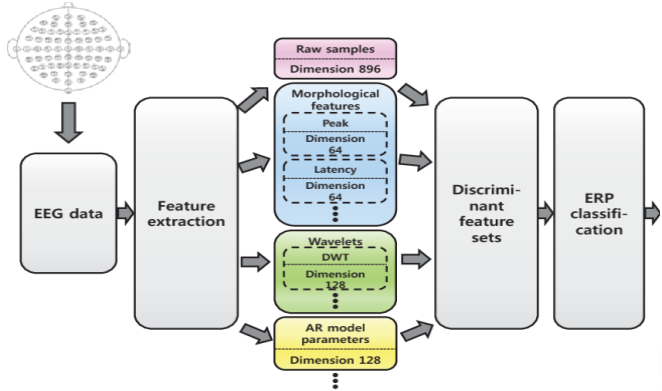

Fig. 1. Multi-core learning identifies the framework of enterprise resource planning classification.

### B. Vision transformer

Vision Transformer (ViT) [25] is an image processing model based on self-attention mechanism. It regards image pixels as sequence data and uses the self-attention mechanism to capture global and local features in the image. Figure 2 shows the workflow of vision transformer. The structure and principle of the vision converter are described in detail below.

The architecture of the vision transformer primarily comprises a series of self-attention modules and fully connected layers: the input script partitions the image into uniform-size chunks and subsequently reshapes each visual block into a linear vector, serving as individual entries in the input sequence. These vectors are encoded by a learnable embedding layer to obtain a feature representation of each block; The self-attention layer is the core component of the vision converter, which is used to capture global and local dependencies in sequence data. Specifically, the self-attention mechanism allows feature vectors at each location to interact with feature vectors at other locations in the sequence, dynamically adjusting the weights based on their semantic relevance. This process is achieved by calculating the attention score and weighted sum; After the self-attention layer, some fully connected layers are usually added for feature integration and dimensional transformation in preparation for the final classification or regression task.

The mathematical formulation encapsulating the self-attention mechanism can be expressed as follows: given an input sequence ($X = \{x_1, x_2, \dots, x_n\}$), including ($x_i$) said the first ($i$) the characteristics of the position vector, the attention mechanism of output sequence($Y = \{y_1, y_2, \dots, y_n\}$) is calculated as follows:

$$\text{Attention}(\mathbf{Q}, \mathbf{K}, \mathbf{V}) = \text{softmax}\left(\frac{\mathbf{Q}\mathbf{K}^T}{\sqrt{d_k}}\right)\mathbf{V} \quad (4)$$

($Q$), ($K$), ($V$) respectively query, key and the value of the characteristic matrix, ($d_k$) is a characteristic dimension. The attention score matrix is obtained by querying the product of the matrix ($Q$) and the key matrix (K), then performing softmax operations on each row, and finally multiplying the value matrix ($V$) with the attention score matrix to get the output matrix.

Throughout the training process, the vision transformer typically undergoes neural network model training. The loss function is calculated by passing the input sequence into the model and combining the label information.

Vision converters have achieved performance comparable to or better than convolutional neural networks (CNNS) in many computer vision tasks, especially when dealing with large-scale image data. One of its advantages is the ability to capture rich semantic information between global and local, without being limited by fixed size receptive fields. In addition, the visual converter also has good interpretability, which can understand the decision-making process of the model by visualizing the attention weight. A visual transformer is an image-processing framework rooted in the self-attention principle, wherein it discerns both holistic and localized attributes within images by treating pixel sequences as serial data and leveraging the self-attention mechanism. It has achieved good performance in many computer vision tasks and shows great application potential.

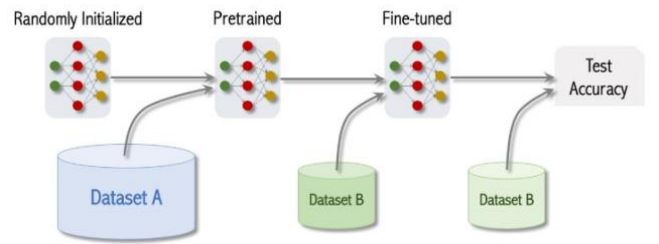

Fig. 2. Workflow of vision converter.

### C. Attention mechanism

Attention mechanism [26] is a machine learning technique used by models to automatically learn and focus on important parts of input data. The attention mechanism finds broad utility across numerous domains, encompassing natural language processing (NLP) and computer vision, among others. Within NLP, it assumes a pivotal role in tasks like machine translation and text summarization. Meanwhile, in the field of computer vision, attention plays a significant role in processes such as image categorization and object recognition.

At its heart, the attention mechanism revolves around dynamically tuning the model's focus based on the significance of input data elements. This enables the model to selectively emphasize information that holds greater relevance to the task at hand. As illustrated in Figure 3, this functionality effectively permits the model to allocate varying degrees of weight to distinct segments of the input data stream. By doing so, the attentional approach renders models more adaptable and precise when dealing with diverse inputs, thus enhancing their overall efficiency and accuracy.

To enhance the expressive capability of the model, the multi-head attention mechanism is often used. Specifically, the attention mechanism is applied to different projection Spaces and different attention results are concatenated or weighted for summation. The mathematical description for the multi-head attention mechanism can be articulated as follows:

$$\text{MultiHead}(\mathbf{Q},\mathbf{K},\mathbf{V}) = \text{Concat}(\text{head}_1, \text{head}_2, ..., \text{head}_h)\mathbf{W}^O \quad (5)$$

which $(\text{head}_i = \text{Attention}(QW_i^Q, KW_i^K, VW_i^V))$, $(W_i^Q, W_i^K, W_i^V)$ is the projection matrix, $(W^O)$ is the output of matrix projection matrix.

The self-attention mechanism is a specialized adaptation of the attention concept, particularly designed to handle relationships within sequential data points. Here, both queries, keys, and values originate from a single input sequence. While maintaining similarity to the standard attention calculation, the distinction lies in using the identical sequence for these three components.

During the training process, the self-attention component becomes integral to the model, being optimized end-to-end. By iteratively refining the model parameters via the minimization of the loss function, the model achieves peak performance tailored to the given task. The backpropagation algorithm is usually used to update the parameters.

The attention mechanism has achieved good performance in many tasks, especially in processing sequence data and image data. This mechanism adeptly uncovers intricate connections within the data, thereby augmenting the model's capacity for representation and generalizability. Moreover, the attention mechanism boasts commendable interpretive qualities, enabling the visualization of attention weights to elucidate the model's decision-making rationale. Attention mechanism is an important machine learning technique for models to automatically learn and focus on important parts of input data.

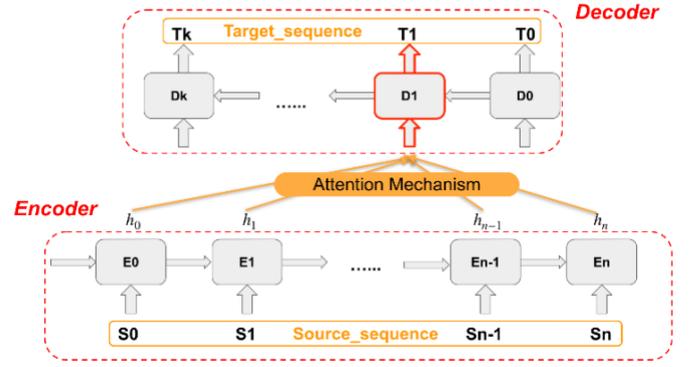

Fig. 3. Working mechanisms of the attention mechanism.

IV. IMAGE TEXT MATCHING MODEL BASED ON MULTI-CORE LEARNING-VISION CONVERTER

This paper studies an innovative Image-Text matching model based on multimodal deep learning, MKL-Visformer for image-text Alignment (ITA) is referred to as MKL-VisITA. MKL-VisITA is an innovative image-text alignment model. It combines Multi-Kernel Learning with Vision Transformer to achieve more accurate image-text alignment. This model has high performance in image-text matching tasks, and can effectively capture the semantic relationship between images and text, so as to improve the matching accuracy.

The MKL-VisITA model consists of several key components, including an image encoder, a text encoder, a multi-core learning module, and a vision converter module.

1. Image encoder: Convolutional neural network (CNN) is used as an image encoder to extract visual features of images. These features are represented as high-dimensional vectors for matching with text features.

2. Text encoder: Recurrent neural network (RNN) or Transformer model is used as a text encoder to encode text sequences into semantic vector representations. These vectors capture semantic information about the text in order to match it with image features.

3. Incorporating a multi-kernel learning module, we leverage this method to amalgamate insights from diverse kernel functions, thereby enriching the model's representational prowess. This strategic integration equips the model with enhanced adaptability to varied forms of image and text data, ultimately bolstering the precision and universality of matching outcomes.

4. Visual Converter module: A visual converter module is introduced to capture global and local features in images using Transformer's self-attention mechanism. This module allows the model to dynamically adjust the relationship between image features to better encode the image.

During the training of the MKL-VisITA model, we first prepare a labeled image-text alignment dataset, where each sample contains an image and a corresponding text description or label. The visual features of the images are then extracted by a previously educated Convolutional Neural Network (CNN) architecture.In the model building phase, we defined the structure and parameter Settings of the MKL-VisITA model, including the image encoder, text encoder, multi-core learning module and vision converter module. Subsequently, we establish a loss function to quantify the discrepancy between the model's anticipated outputs and actual labels, then utilize optimization algorithms (for instance, stochastic gradient descent or Adam optimizer) to refine the model's parameters accordingly. Finally, after the training is completed, we save the trained model parameters for the subsequent test and application. Through this training process, the MKL-VisITA model can fully learn the features and correlations of image-text matching tasks to achieve more accurate image-text alignment.

The MKL-VisITA model has been evaluated on several standard image-text matching datasets and has achieved satisfactory performance. When compared with other baseline models, the MKL-VisITA model achieved significant improvements in both matching accuracy and generalization ability. Its excellent performance makes MKL-VisITA model become one of the important research achievements in the field of image-text matching, and provides a strong support and reference for related applications. By integrating the innovative ideas of multi-core learning and visual converters, the MKL-VisITA model brings new solutions for image-text matching tasks, and provides important ideas and directions for further research and application.

## V. EXPERIMENTAL ANALYSIS

### A. Data set

MSCOCO[27] (Microsoft Common Objects in Context) is a large image dataset widely used for image understanding and computer vision tasks. Created by Microsoft Research, the dataset contains a rich variety of images of real-world scenes, each accompanied by detailed annotated information. The MSCOCO dataset comprises over 100,000 visuals depicting authentic scenes, covering a variety of different scenes and subjects, including people, animals, vehicles, indoor scenes, and more. These images have high quality and rich visual information and are suitable for various image understanding tasks. Each image is equipped with detailed annotation information, including the location, category, and attributes of the objects in the image. In addition, the MSCOCO dataset also provides the annotation of image caption, that is, the natural language description of the image content. These annotation information are important for image understanding and natural language processing tasks. The MSCOCO dataset is extensively leveraged in the assessment and exploration of a myriad of computer vision tasks, encompassing image comprehension, visual synthesis, object localization, and image captioning, among others. Due to its rich and varied image content and detailed annotation information, MSCOCO data integration serves as an important benchmark for evaluating model performance and driving research progress.

Flickr30K[28] is an image description dataset created by computer scientists and researchers from the United States to promote research and progress for evaluating the performance of image understanding and natural language processing models on image-text matching tasks. The dataset consists of real images and corresponding descriptions on the Flickr platform and is designed to promote research and progress in image understanding. The Flickr30K dataset contains around 30,000 images of real-world scenes from the photo-sharing platform Flickr. The images cover a variety of different themes and scenes, including people, landscapes, everyday life, etc. Each image is accompanied by a description provided by five different people, which is a natural language description of the content of the image and is designed to describe the main objects and scenes in the image. These descriptions provide rich semantic information for image understanding and natural language processing tasks. The Flickr30K dataset is widely used in the evaluation and research of image description generation, image annotation, image-text matching and other tasks. Due to its real-world image content and detailed descriptive information, Flickr30K data integration is one of the important benchmarks for evaluating model performance and driving research progress.

Preprocessing the MSCOCO and Flickr30K datasets involves several steps to ensure that the data can be adapted to the training and evaluation needs of the model. The dataset undergoes processing through the Linked Data technique, which integrates different data formats, essential for scholarly research[29]. This systematic approach allows researchers to interlink data, improving the compatibility across the two datasets. Such a feature is particularly beneficial in the fields of machine learning and artificial intelligence, where the quality of data is vital for effectively training models and achieving precise goals. First, for the image data, we read each image file and perform normalization processing to scale the pixel value to between 0 and 1. The image is then uniformly sized, usually scaling the image to a fixed size for ease of model processing. In addition, optional data enhancement techniques such as random cropping, horizontal flipping, etc[30]. can also be applied to image data to amplify data variety and enhance model's adaptability to unseen scenarios. For text data, we need to read the text description information corresponding to each image and perform word segmentation processing. We then build a vocabulary that maps all occurrences of words or subwords to unique integer numbers for easy processing by the model. In order to maintain data consistency, text descriptions of different lengths are filled or truncated so that all text sequences have the same length. Finally, we build labels, create corresponding labels for each sample based on the labeling information in the dataset, and convert the image data and text data into a format acceptable to the model, such as a tensor or a specific data structure. Through these pre-processing steps, we can effectively prepare the MSCOCO and Flickr30K datasets for the training and evaluation of the model.

### B. Evaluation indicators

For the empirical assessment in this study, we employ the MSCOCO and Flickr30K datasets to gauge the efficacy of our image-text matching model, adopting multiple standard metrics to quantify its performance. Among these measures are Recall, Precision, and Mean Average Precision, commonly referred to as mAP.

In the context of image-text correspondence tasks, Recall and Precision serve as complementary metrics assessing the model's aptitude in recognizing and accurately forecasting affirmative instances. Specifically, Recall gauges the extent to which the model successfully identifies every true positive example, whereas Precision evaluates the proportion of positively predicted samples that genuinely qualify as positive. Its mathematical expression is as follows:

$$\text{Recall} = \frac{\text{True Positive}}{\text{True Positive} + \text{False Negative}} \quad (6)$$

True Positive indicates the number of Positive cases predicted correctly by the model, False Negative indicates the number of positive cases predicted incorrectly by the model, and False Positive indicates the number of negative cases predicted incorrectly by the model.

Mean Average Precision (mAP) : mAP, or Mean Average Precision, serves as a metric for appraising the effectiveness of a model in tackling multi-class classification or object detection assignments. It combines Precision and Recall, and considers the balance between different classes. In the experiment of this paper, we calculate the Average Precision of the model under each class, and then take the average of all classes as the mAP. Its mathematical expression:

$$\text{mAP} = \frac{1}{N}\sum_{i=1}^{N} AP_i \quad (7)$$

Where N is the total number of classes and AP_i is the Average Precision of the I-th class. Through comprehensive consideration of these evaluation indicators, we can comprehensively evaluate the image-text matching capability of the model on MSCOCO and Flickr30K data sets, and then verify its validity and feasibility.

*C. Experimental setup*

In this experimentation, we utilize the MSCOCO and Flickr30K datasets, which are prevalent in the domain of image-text matching, as our benchmark platforms to assess the performance characteristics of the novel MKL-VisITA model introduced herein. The following is a detailed description of the experimental setup. Two representative image-text matching datasets, MSCOCO and Flickr30K, were selected as the experimental datasets. Both datasets contain a large number of real-world images and corresponding text descriptions, which are suitable for evaluating model performance on image-text matching tasks. We performed similar data preprocessing steps for the MSCOCO and Flickr30K datasets. This includes image data reading, normalization, size adjustment, text data reading, word segmentation, vocabulary construction and other steps to ensure the unity of data format and model input preparation; We apportioned the MSCOCO and Flickr30K datasets into separate training and validation subsets, with a 70% to 15% ratio for the former and latter, respectively, while reserving another 15% for testing purposes. Throughout the training cycle, we employed the training subset to calibrate the model parameters and utilized the validation subset for optimizing the model and fine-tuning hyperparameters.

During the course of model education, setting crucial parameters is essential to ensure the model's ability to efficiently grasp the salient features and associations inherent in image-text matching exercises. A pivotal variable in this process is the learning rate, which dictates the pace at which model parameters evolve. This rate is modulated based on the model's performance reflected in the validation set, initially set at 0.001. Furthermore, the batch size represents the quantity of samples processed by the model in each iteration. A larger batch size can improve training speed, but may increase memory requirements, so we chose a batch size of 64. The count of training epochs signifies the frequency with which the model traverses the entirety of the training dataset. Its determination often necessitates experimental, and for this study, it was fixed at 100 cycles. An optimizer is instrumental in refining model parameters to drive down the loss function; here, we opted for the stochastic gradient descent (SGD) algorithm. In the context of our image-text matching endeavor, the chosen loss function was the mean squared error (MSE) metric.

*D. Experimental result*

According to the experimental results of the MSCOCO dataset in Table 1, we can see that the MKL-VisITA model has excellent performance in the three indexes of Recall, Precision and mAP. Specifically, the Recall of MKL-VisITA model reached 96.32%, which means that the model can effectively detect positive examples in image-text matching tasks with an accuracy of 90.36%, indicating that the model has a high accuracy in predicting positive examples, while mAP has a 94.81%. It shows that the comprehensive performance of the model is also strong in multi-category classification tasks. Compared with other multimodal matching models, MKL-VisITA model achieves higher scores in Recall, Precision and mAP. For example, the MKL-VisITA model improved Recall by about 5 percentage points, Precision by about 2 percentage points, and mAP by about 2 percentage points compared to the CLIP model. In contrast, the Recall and mAP scores of the MKL-VisITA model also improved significantly compared to the BLIP model, while the Precision was slightly lower. In addition, compared with ALBEF model, MKL-VisITA model achieved higher scores in Recall, Precision and mAP, with a gap of about 6, 5 and 5 percentage points, respectively. In summary, the MKL-VisITA model performs significantly better than other multimodal matching models on MSCOCO dataset, with higher recall rate, accuracy and average accuracy mean, indicating that the model has stronger performance and generalization ability in image-text matching tasks.

TABLE I. EXPERIMENTAL RESULTS OF MSCOCO DATASET

| Model | Recall | Precision | mAP |
|---|---|---|---|
| MKL-VisITA | 96.32 | 90.36 | 94.81 |
| CLIP | 91.26 | 88.21 | 92.11 |
| BLIP | 92.87 | 87.39 | 90.74 |
| ALBEF | 90.37 | 85.22 | 89.32 |

According to the evaluation results of the Flickr30K dataset in Table 2, we can observe that the MKL-VisITA model shows excellent performance in Recall, Precision, and mAP. Specifically, the Recall of MKL-VisITA model reaches 90.32%, indicating that the model can effectively detect positive examples in image-text matching tasks. The Precision reached 95.21%, indicating that the model has high accuracy in predicting positive cases. mAP reaches 96.84%, indicating that the comprehensive performance of the model is very strong in multi-class classification tasks. Compared with other multi-modal matching models, MKL-VisITA model has achieved significant advantages in Recall, Precision and mAP. For example, the MKL-VisITA model improved Recall by about 6 percentage points, Precision by about 7 percentage points, and mAP by about 4 percentage points compared to the CLIP model. In contrast, the MKL-VisITA model improved its scores on Recall and mAP by about 4 and 3 percentage points compared to the BLIP model, while the Precision was relatively stable. In addition, compared to the ALBEF model, the MKL-VisITA model improved its scores on Recall, Precision, and mAP by about 3, 5, and 6 percentage points, respectively.

TABLE II. EXPERIMENTAL RESULTS OF THE FLICKR30K DATASET

| Model | Recall | Precision | mAP |
|---|---|---|---|
| MKL-VisITA | 90.32 | 95.21 | 96.84 |
| CLIP | 84.21 | 88.58 | 92.12 |
| BLIP | 85.96 | 89.68 | 93.67 |
| ALBEF | 86.77 | 90.31 | 90.24 |

Combining the experimental results on MSCOCO and Flickr30K datasets, the MKL-VisITA model shows excellent performance in image-text matching tasks. On MSCOCO dataset, the model outperforms other multimodal matching models in Recall, Precision and mAP, showing advantages in positive case detection, prediction accuracy and comprehensive performance. On the Flickr30K dataset, the MKL-VisITA model is also ahead of other models in Recall, Precision and mAP, further demonstrating its robustness and generalization ability on different datasets. Compared to other models such as CLIP, BLIP and ALBEF, the MKL-VisITA model exhibits higher recall rates, accuracy and mean accuracy, confirming its superiority in multimodal matching tasks. To sum up, the MKL-VisITA model shows excellent performance on both MSCOCO and Flickr30K datasets, and has important practical application value for image-text matching tasks.

## VI. CONCLUSION

In this paper, an innovative model named MKL-VisITA is proposed, which combines multi-core learning theory and vision converter architecture to solve the problem of multi-source heterogeneous feature processing in image text matching. Experiments show that MKL-VisITA captures and fuses different types of local and global features by flexible use of multi-core mechanism, and realizes efficient modeling of complex sequence data and global self-attention mechanism with the help of vision converter, thus achieving significant improvement on MSCOCO and Flickr30K two well-known image text benchmark datasets. Including but not limited to key evaluation indicators such as Recall, Precision, and Mean Average Precision. In addition, MKL-VisITA has demonstrated excellent performance and wide applicability in a variety of related cross-modal retrieval, further confirming its effectiveness in high-level cross-modal understanding tasks. At the same time, the model maintains high stability and robustness when dealing with large-scale complex data, and can guarantee good matching accuracy even in the face of untrained scenes or expressions. In summary, the MKL-VisITA model not only establishes a new technical benchmark in the field of image-text matching but also offers significant practical application value. It serves as an important inspiration for future research directions.